\documentclass{article} 
\usepackage{iclr2020_conference,times}

\usepackage{amsmath,amsfonts,bm}

\def\eqref#1{equation~\ref{#1}}

\def\1{\bm{1}}

\def\ru{{\textnormal{u}}}

\def\rw{{\textnormal{w}}}
\def\rx{{\textnormal{x}}}

\def\rz{{\textnormal{z}}}

\def\vtheta{{\bm{\theta}}}

\def\vb{{\bm{b}}}
\def\vc{{\bm{c}}}

\def\mA{{\bm{A}}}

\def\mL{{\bm{L}}}

\def\mQ{{\bm{Q}}}

\def\mT{{\bm{T}}}

\DeclareMathAlphabet{\mathsfit}{\encodingdefault}{\sfdefault}{m}{sl}
\SetMathAlphabet{\mathsfit}{bold}{\encodingdefault}{\sfdefault}{bx}{n}

\newcommand{\E}{\mathbb{E}}

\newcommand{\R}{\mathbb{R}}

\usepackage{hyperref}
\usepackage{url}
\usepackage{graphicx}
\graphicspath{ {./figures/} }
\usepackage{subfig}
\usepackage{makecell}
\usepackage{enumitem}

\newtheorem{theorem}{Theorem}

\title{Disentanglement by Nonlinear ICA with \\ General Incompressible-flow Networks (GIN)}

\author{Peter Sorrenson, Carsten Rother, Ullrich K\"othe \\
Visual Learning Lab\\
Heidelberg University\\
}

\iclrfinalcopy 
\begin{document}

\maketitle

\begin{abstract}
A central question of representation learning asks under which conditions it is possible to reconstruct the true latent variables of an arbitrarily complex generative process. Recent breakthrough work by \citet{khemakhem2019variational} on nonlinear ICA has answered this question for a broad class of \emph{conditional} generative processes. We extend this important result in a direction relevant for application to real-world data. First, we generalize the theory to the case of unknown intrinsic problem dimension and prove that in some special (but not very restrictive) cases, informative latent variables will be automatically separated from noise by an estimating model. Furthermore, the recovered informative latent variables will be in one-to-one correspondence with the true latent variables of the generating process, up to a trivial component-wise transformation. Second, we introduce a modification of the RealNVP invertible neural network architecture \citep{dinh2016density} which is particularly suitable for this type of problem: the General Incompressible-flow Network (GIN). Experiments on artificial data and EMNIST demonstrate that theoretical predictions are indeed verified in practice. 
In particular, we provide a detailed set of exactly 22 informative latent variables extracted from EMNIST.
\end{abstract}


\section{Introduction}

Deep latent-variable models promise to unlock the key factors of variation within a dataset, opening a window to interpretation and granting the power to manipulate data in an intuitive fashion. The theory of identifiability in linear independent component analysis (ICA) \citep{comon1994independent} tells us when this is possible, if we restrict the model to a linear transformation, but until recently there was no corresponding theory for the highly nonlinear models needed to manipulate complex data.
This changed with the recent breakthrough work by \citet{khemakhem2019variational}, which showed that under relatively mild conditions, it is possible to recover the joint data and latent space distribution, up to a simple transformation in the latent space. The key requirement is that the generating process is conditioned on a variable which is observed along with the data. This condition could be a class label, time index of a time series, or any other piece of information additional to the data. They interpret their theory as a \emph{nonlinear} version of ICA.

This work extends this theory in a direction relevant for application to real-world data. The existing theory assumes knowledge of the intrinsic problem dimension, but this is unrealistic for anything but artificially generated datasets. Here, we show that in the special case of Gaussian latent space distributions, the intrinsic problem dimension can be \emph{discovered}. The important latent variables are organically separated from noise variables by the estimating model. Furthermore, the variables discovered correspond to the true generating latent variables, up to a trivial component-wise translation and scaling. Very similar results exist for other members of the exponential family with two parameters, such as the beta and gamma distributions.

We introduce a variant of the RealNVP \citep{dinh2016density} invertible neural network: the General Incompressible-flow Network (GIN). The flow is called \emph{incompressible} in reference to fluid dynamics, since it preserves volumes: the Jacobian determinant is simply unity. We emphasise its \emph{generality} and increased expressive power in comparison to previous volume-preserving flows, such as NICE \citep{dinh2014nice}.
As already noted in \citet{khemakhem2019variational}, flow-based generative models are a natural fit for the theory of nonlinear ICA, as are the variational autoencoders (VAEs) \citep{kingma2013auto} used in that work. For us, major advantages of invertible architectures over VAEs are the ability to specify volume preservation and directly optimize the likelihood, and freedom from the requirement to specify the dimension of the model's latent space. An INN always has a latent space of the same dimension as the data. In addition, the forward and backward models share parameters, saving the effort of learning separate models for each direction.

In summary, our work makes the following contributions:
\begin{itemize}[itemsep=2pt,parsep=2pt,topsep=0pt, partopsep=0pt]
    \item We extend the theory of nonlinear ICA to allow for unknown intrinsic problem dimension. Doing so, we find that this dimension can be discovered and a one-to-one correspondence between generating and estimated latent variables established.
    \item We propose as an implementation an invertible neural network obtained by modifying the RealNVP architecture.
    We call our new architecture GIN: the General Incompressible-flow Network.
    \item We demonstrate the viability of the model on artificial data and the EMNIST dataset. We extract 22 meaningful variables from EMNIST, encoding both global and local features.
    \vspace{-2mm}
\end{itemize}

\section{Related Work}
The basic goals of nonlinear ICA stem from the original work on linear ICA. An influential formulation, as well as the first identifiability results, were given in \citet{comon1994independent}. These stated the conditions which allow the generating latent variables to be discovered, when the mixing function is a linear transformation. However, it was shown in \citet{hyvarinen1999nonlinear} that this approach to identifiability does not extend to general nonlinear functions.

The first identifiability results in nonlinear ICA came in \citet{hyvarinen2016unsupervised} and \citet{hyvarinen2017nonlinear}, applied to time series, and implemented via a discriminative model and semi-supervised learning. A more general formulation, valid for other forms of data, was given in \citet{hyvarinen2018nonlinear} and the theory was extended to generative models in \citet{khemakhem2019variational}, where experiments were implemented by a VAE.

Many authors have addressed the general problem of \emph{disentanglement}, and proposed models to learn disentangled features. Prominent among these is $\beta$-VAE \citep{higgins2017beta} and its variations \citep[e.g.][]{chen2018isolating} which augment the standard ELBO loss with tunable hyperparameters to encourage disentanglement. There are also attempts to modify the GAN framework \citep{goodfellow2014generative} such as InfoGAN \citep{chen2016infogan}, which tries to maximize the mutual information between some dimensions of the latent space and the observed data. Many of these approaches are unsupervised. However, as pointed out and empirically demonstrated in \citet{locatello2018challenging}, unsupervised models without conditioning in the latent space are in general unidentifiable.

Several unsupervised VAE models implement conditioning in the latent space by means of Gaussian mixtures \citep{johnson2016composing,dilokthanakul2016deep,zhao2019variational}. Our work differs mainly by (i) only considering supervised tasks, therefore being safely covered by the theory of \citet{khemakhem2019variational}, and (ii) enforcing volume-preservation, not possible in a VAE.

The invertible neural networks in this work build upon the NICE framework \citep{dinh2014nice} and its extension in RealNVP \citep{dinh2016density}. A similar network design to ours is \citet{ardizzone2019guided}. This is a conditioned INN based on RealNVP, however the conditioning information is applied as a parameter of the network. 
The authors find in experiments with MNIST that meaningful variables are present in their latent space, but are rotated such that they are not aligned with the axes of the space. In this work, the conditioning is only present as a parameter of the latent space distributions. As a result, it is covered by the theory of \citet{khemakhem2019variational} and its extension here, which results in non-rotated, meaningful latent variables.

\section{Theory}
\subsection{Existing Theory of Nonlinear ICA} \label{existing theory section}

\begin{figure}[t] 
    \vspace{-12mm}
    \begin{center}
        \includegraphics[width=\textwidth]{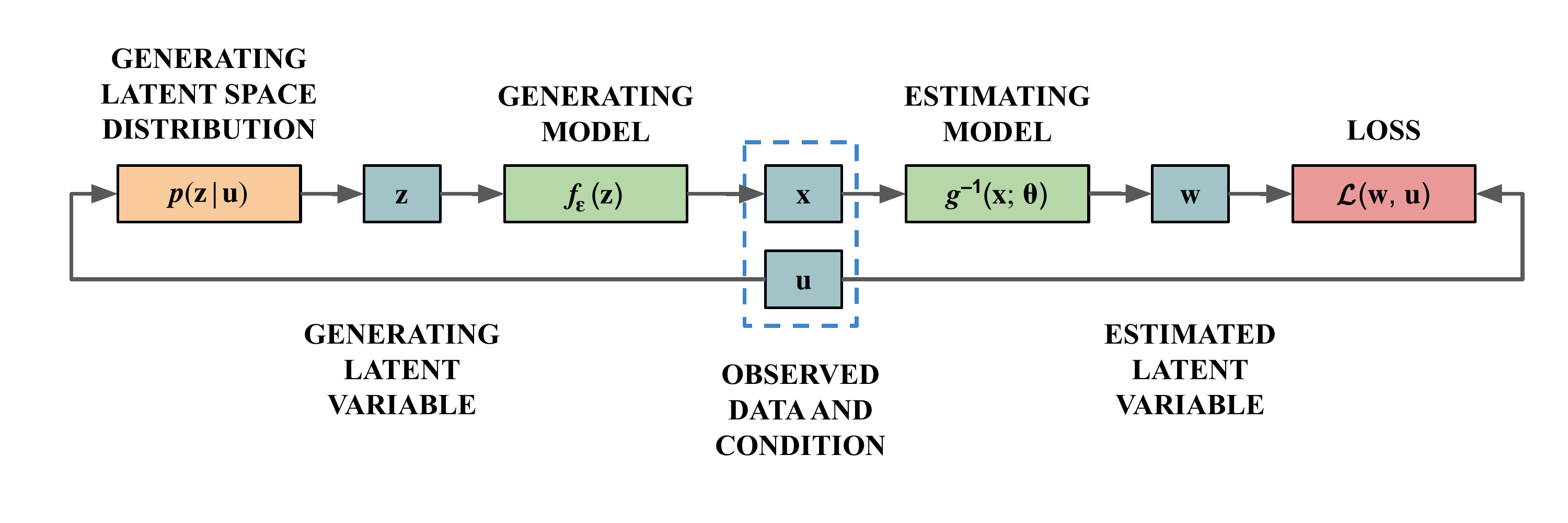}
    \end{center}
    \vspace{-6mm}
    \caption{Relationship between the variables and functions defined in Sec.~\ref{existing theory section}, as well as indication of the training scheme.}
\end{figure}

This section adapts theoretical results from \citet{khemakhem2019variational} to the context of invertible neural networks.
Suppose the existence of the following three random variables: a latent generating variable $\rz \in \R^n$, a condition $\ru \in \R^m$ and a data point $\rx \in \R^d$, where $n \leq d$. If $n < d$, also suppose the existence of a noise variable $\varepsilon \in \R^{d-n}$. The variables $\rz$ and $\varepsilon$ make up the generating latent space, and are not necessarily known, whereas $\ru$ and $\rx$ are observable and therefore always known. 

The distribution of $\rz$ is a factorial member of the exponential family with $k$ sufficient statistics, conditioned on $\ru$. In its most general form the distribution can be written as
\begin{equation} \label{eqn:z latent space}
    p_{\rz}(\rz | \ru) = \prod_{i=1}^n \frac{Q_i(z_i)}{Z_i(\ru)} \exp \left[ \sum_{j=1}^k T_{i,j}(z_i) \lambda_{i,j}(\ru) \right]
\end{equation}
where the $T_{i,j}$ are the sufficient statistics, the $\lambda_{i,j}$ their coefficients and $Z_i$ the normalizing constant. $Q_i$ is called the base measure, which in many cases is simply 1. 

The distribution of $\varepsilon$ must not depend on $\ru$ and its probability density function must be always finite. 

The variable $\rx$ is the result of an arbitrarily complex, invertible, deterministic transformation from the generating latent space to the data space: $\rx = f(\rz, \varepsilon)$. This can alternatively be formulated as an injective, non-deterministic transformation from the lower-dimensional $\rz$-space to the higher-dimensional $\rx$-space: $\rx = f_{\varepsilon}(\rz)$. 

In general, an observed dataset $\mathcal{D}$ will consist only of instances of $\rx$ and $\ru$. The task of nonlinear ICA is to \emph{disentangle} the data to recover the generating latent variables $\rz$, as well as the form of the function $f$ and its inverse. We can try to achieve this with a sufficiently general, invertible function approximator $g$, which maps from a latent variable $\rw \in \R^d$ to the data space: $\rx = g(\rw; \theta)$. Here $\theta$ denotes the parameters of $g$. Note that the dimensions of the latent space and data space are the same due to the invertibility of $g$. We assume that $\rw$ follows a conditionally independent exponential probability distribution, conditioned on the same condition $\ru$ as $\rz$:
\begin{equation} \label{eqn:w latent space}
    p_{\rw}(\rw | \ru) = \prod_{i=1}^d \frac{Q'_i(w_i)}{Z'_i(\ru)} \exp \left[ \sum_{j=1}^k T'_{i,j}(w_i) \lambda'_{i,j}(\ru) \right].
\end{equation}
The coefficients of the sufficient statistics $\lambda_{i,j}'(\ru)$ are not restricted, which means that a given variable $i$ of the estimated latent space is allowed to lose its dependence on $\ru$. If this occurs, we will consider this variable to encode noise, playing a role equivalent to $\varepsilon$ in the generating latent space.

In addition to this specification of the generative and estimating models, some conditions are necessary to ensure the latent variables can be recovered. The most important of these concerns the variability of $\lambda_{i,j}(\ru)$ under $\ru$. However, as long as the $\lambda_{i,j}$ are randomly and independently generated, and there are at least $nk+1$ distinct conditions $\ru$, this condition is almost surely fulfilled. See Appendix \ref{appendix:remark} for further details.

In the limit of infinite data and perfect convergence, the estimating model will give the same conditional likelihood to all data points as the true generating model: 
\begin{equation} \label{eqn:conditional likelihood x}
    p_{\mathbf{T},\bm{\lambda},f,\varepsilon}(\rx|\ru) = p_{\mathbf{T}',\bm{\lambda}',g}(\rx|\ru).
\end{equation}
If this is the case, the vector of sufficient statistics $\mathbf{T}$ from the generating latent space will be related to that of the estimated latent space by an affine transformation:
\begin{equation} \label{latent space relation}
    \mathbf{T}(\rz) = \mA \mathbf{T}'(\rw) + \vc,
\end{equation} 
where $\mA$ is some constant, full-rank, $nk \times dk$ matrix and $\vc \in \R^{nk}$ some constant vector. The relationship holds for all values of $\rz$ and $\rw$. The proof of this result can be found in Appendix \ref{appendix:proofs}.

This relationship is a generalization of that derived in \citet{khemakhem2019variational}. In that version, they assume knowledge of $n$, the dimension of the generating latent space, and give their estimating model a latent space of the same dimension. In this context the matrix $\mA$ defines a relationship between latent spaces of the same dimension, making it square and invertible.

\subsection{Nonlinear ICA with a Gaussian latent space}
Given the relationship in equation (\ref{latent space relation}), we can ask whether any stronger results hold for particular special cases. We hope in particular to induce sparsity in the matrix $\mA$, so that the estimated latent space is related to the true one by as simple a transformation as possible.
We show (proof in Appendix \ref{gaussian general case}) that when both the generating and estimated latent spaces follow a Gaussian distribution, the generating latent space variables are recovered up to a trivial translation and scaling. Furthermore, the dimension of the generating latent space is recovered.

More precisely, each generating latent variable $z_i$ is related to exactly one estimated latent variable $w_j$, for some $j$, as
$z_i = a_i w_j + b_i$,
for some constant $a_i$ and $b_i$. Furthermore, each estimated latent variable $w_j$ is related to at most one $z_i$. If the estimating latent space has higher dimension than the dimension of the generating latent space, some estimating latent variables are not related to any generating latent variables and so must encode only noise. 
This is the dimension discovery mechanism, since the estimated latent space organically splits into informative and non-informative parts, with the dimension of the informative part equal to the unknown intrinsic dimension of the generating latent space.
Very similar results can be derived for all common continuous two-parameter members of the exponential family, including the gamma and beta distributions. See Appendix \ref{two parameter general case}.

\subsection{Volume Preservation via Incompressible Flow}
In a multivariate Gaussian distribution with diagonal covariance, the standard deviations are directly proportional to the principal axes of the ellipsoid defining a surface of constant probability. In classical PCA, this property is used to assign importance to each principal component based on its standard deviation. In PCA we can think of the rotation of the data into the basis defined by the principal components as the latent space. In a deep latent variable model the transformation between the latent space and the data is significantly more complex, but if this transformation preserves volumes like the rotation in PCA, it will retain the desirable correspondence between the standard deviation of a latent space variable and its importance in explaining the data. Because invertible neural networks (normalizing flows) have a tractable Jacobian determinant, they open the possibility to constrain it to unity. This is equivalent to volume preservation and is the rationale behind the General Incompressible-flow Network.

\section{Experiments} \label{section:experiments}
Experiments on artificial datasets confirm the theory for a normally distributed latent space, as well as identifying potential causes of failure. Experiments on the EMNIST dataset \citep{cohen2017emnist} demonstrate the ability of GIN to estimate independent and interpretable latent variables from real-world data.

\subsection{Model Description}
The GIN model is similar in form to RealNVP \citep{dinh2016density} and shares its flexibility, but retains the volume-preserving properties of the NICE framework \citep{dinh2014nice}. 

RealNVP coupling layers split $D$-dimensional input $x$ into two parts, $x_{1:d}$ and $x_{d+1:D}$ where $d < D$. The output of the layer is the concatentation of $y_{1:d}$ and $y_{d+1:D}$ with
\begin{align}
    y_{1:d} &= x_{1:d} \\
    y_{d+1:D} &= x_{d+1:D} \odot \exp(s(x_{1:d})) + t(x_{1:d})
\end{align}
where addition, multiplication and exponentiation are applied component-wise. The logarithm of the Jacobian determinant of a coupling layer is simply the sum of the scaling function $s(x_{1:d})$. Volume preservation is achieved by setting the final component $s(x_d)$ to the negative sum of the previous components, so the total sum of $s(x_{1:d})$ is zero (in contrast, NICE sets $s(x_{1:d})\equiv 0$). Hence the Jacobian determinant is unity and volume is preserved. The network is free to allow the volume contribution of some dimensions of the output of any coupling layer to grow, but only by shrinking the other dimensions of the output in direct proportion. As well as enforcing a strong correspondence between the importance of a latent variable and its standard deviation, we believe volume-preservation has a regularizing effect, since it is a very strong constraint, comparable to orthonormality in linear transformations.

\subsection{Optimization} \label{section:optimization}
The experiments in this section deal with labeled data, where each data point belongs to one of $M$ different classes. This class label is used as the condition $\ru$ associated with each data point $\rx$. In the estimated latent space, all data instances with the same label should belong to the same Gaussian distribution. Hence we are learning a Gaussian mixture in the estimated latent space, with $M$ mixture components. Since the distribution for each class in the estimated latent space is required to be factorial, the variance of each mixture component is diagonal, and we can write $\sigma_i^2(\ru)$ for the variance in the $i$-th dimension.

Given a set of data, condition pairs $\mathcal{D} = \{(\rx^{(1)},\ru^{(1)}), \dots, (\rx^{(N)},\ru^{(N)})\}$ and model $g$, parameterized by $\vtheta$ (where $g$ maps from the latent space to the data space) we can construct a loss from the log-likelihood. Using the change of variables formula, we have $\log p(\rx|\ru) = \log p(\rw|\ru)$, with no Jacobian term since the transformation $\rw = g^{-1}(\rx; \vtheta)$ is volume-preserving. To maximize the likelihood of $\mathcal{D}$, we minimize the negative log-likelihood of $\rw$ in the estimated latent space:
\begin{equation}
    \mathcal{L}(\vtheta) = \E_{(\rx,\ru) \in \mathcal{D}} \left[ \frac{1}{n} \sum_{i=1}^n \left( \frac{(g^{-1}_i(\rx; \vtheta)-\mu_i(\ru; \vtheta))^2}{2\sigma_i^2(\ru; \vtheta)} + \log(\sigma_i(\ru; \vtheta)) \right) \right].
\end{equation}


\subsection{Artificial Data}
\subsubsection{Experiment 1: conditions of theory fulfilled}

\begin{figure}[t] 
    \vspace{-4mm}
    \begin{center}
        \includegraphics[width=0.9\textwidth]{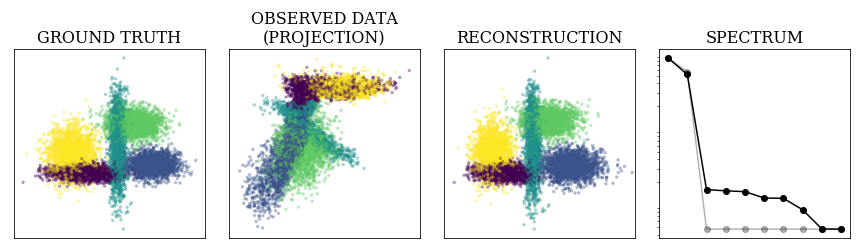}
    \end{center}
    \vspace{-5mm}
    \caption{Successful reconstruction by GIN of the two informative latent variables out of ten in total. The other eight are correctly identified as noise. The observed data is ten-dimensional (projection into two dimensions shown here). The spectrum shows the standard deviation of each variable of the reconstruction (in black, log scale) which quantifies its importance. Ground truth is in gray. There is a clear distinction between the two informative dimensions and the noise dimensions, showing that GIN has correctly detected a two-dimensional manifold in the ten-dimensional data presented to it. 
    \vspace{-4mm}
    }
    \label{teaser fig}
\end{figure}

Samples are generated in two dimensions, conditioned on five different cluster labels, see Fig. \ref{teaser fig}. 
The means of the clusters are chosen independently from a uniform distribution on $[-5, 5]$ and variances from a uniform distribution on $[.5, 3]$.\footnote{These values are the same as in an artificial data experiment in \citet{khemakhem2019variational}. The difference is that the data will be projected into a higher-dimensional space and that we will not assume that the estimating latent space is two-dimensional, this will instead be inferred by the model.} 
This data is then concatenated with independent Gaussian noise in eight dimensions to make a ten-dimensional generating latent space, where only the first two variables are informative. The noise is scaled by 0.01 to be small in comparison to the informative dimensions. The latent space samples are then passed through a RealNVP network with 8 fully connected coupling blocks with randomly initialized weights to produce the observed data. This acts as a highly nonlinear mixing which can only be successfully treated with nonlinear methods. 

GIN is used as the estimating model, with 8 fully connected coupling blocks (full details in Appendix \ref{inn architecture}). Training converges quickly and stably using the Adam optimizer \citep{kingma2014adam} with initial learning rate $10^{-2}$ and other values set to the usual recommendations. Batch size is 1,000 and the data is augmented with Gaussian noise ($\sigma = 0.01$) at each iteration. After convergence of the loss, the learning rate is reduced by a factor of 10 and trained again until convergence.

Over a number of experiments we made the following observations:
\begin{itemize}[itemsep=2pt,parsep=2pt,topsep=0pt, partopsep=0pt]
    \item The model converges stably and gives importance (quantified by standard deviation) to only two variables in the estimated latent space, provided there is sufficient overlap between the mixture components in the generating latent space.
    \item Where there is not enough overlap in the generating latent space, the model cannot recognize common variables across all the different classes, and tends to split one genuine dimension of variation into two or more in its estimated latent space. 
    This appears to be a problem of finite data. We have observed that when this behaviour occurs, and if the gap between mixture components is not too large, it can be prevented by increasing the number of samples so that the space between the mixture components is better filled (see Fig. \ref{5 clusters large dataset fig} and \ref{5 clusters small dataset fig} in Appendix \ref{appendix:artificial figs}). This is consistent with the theory, where equation (\ref{latent space relation}) is true asymptotically. Since the latent space distributions are members of the exponential family, they have support across the entire domain of the latent space, hence gaps can never remain in the limit of infinite samples.
    \item Choice of learning rate is important. If the initial learning rate is too low, training gets stuck in bad local optima, where single true variables are split into several latent dimensions.
\end{itemize}

\subsubsection{Experiment 2: conditions of theory not fulfilled}


Samples are generated as in Experiment 1, but with only three mixture components. Since there are two sufficient statistics per dimension, and two dimensions of variation, according to the theory we need at least $nk+1=5$ distinct conditions $\ru$ for equation (\ref{latent space relation}) to hold (see section \ref{existing theory section}). Therefore, we might not expect successful experiments. Nonetheless, we observe essentially the same results as for the previous experiments (see Fig. \ref{3 clusters fig} in Appendix \ref{appendix:artificial figs}), with the same caveats regarding gaps between the mixture components in the generating latent space. This suggests that the conditions derived in \citet{khemakhem2019variational}, although sufficient for disentanglement, are not necessary.

\subsection{EMNIST}
\subsubsection{Experiment}

\begin{figure}[ht] 
    \vspace{-3mm}
    \begin{center}
        \includegraphics[width=0.85\textwidth]{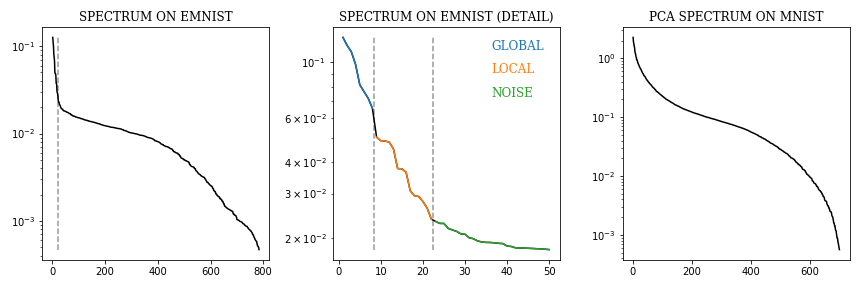}
    \end{center}
    \vspace{-5mm}
    \caption{Spectrum of sorted standard deviations derived from training GIN on EMNIST. On the right is the equivalent spectrum from PCA, a linear method, on MNIST. The nonlinear spectrum exhibits a sharp knee not obtained by linear methods. In the nonlinear spectrum, the first 22 latent variables encode information about the shape of a digit, while the rest of the latent variables encode noise. This distinction is marked with a dotted line in the left and center figures (see Sec.~\ref{sec:emnist results} for explanation of the choice of this cut-off). Within the first 22 variables, the first eight encode global information, such as slant and width, whereas the following 14 encode more local information. This distinction is marked in the center figure only. \vspace{-4mm}}
\end{figure}

The data comes from the EMNIST Digits training set of 240,000 images of handwritten digits with labels \citep{cohen2017emnist}. EMNIST is a larger version of the well-known MNIST dataset, and also includes handwritten letters. Here we use only the digits. The digit label is used as the condition $\ru$, hence we construct a Gaussian mixture in the estimated latent space with 10 mixture components. According to the theory (see \ref{existing theory section}), these are only enough conditions to guarantee identifiability if there are only four informative latent variables in the generating latent space. We expect that the true number of informative variables is somewhat higher, so as in Experiment 2 above, we are operating outside of the guarantees of the theory. In addition, the true generative process of human handwriting may not exactly fulfill our method's assumptions, and we might still lack data on rare or subtle variations, despite the large size of the dataset in comparison to MNIST.

The estimating model is a GIN which uses convolutional coupling blocks and fully connected coupling blocks to transform the data to the latent space (full details in Appendix \ref{inn architecture}). Optimization is with the Adam optimizer, with initial learning rate 3e-4. Batch size is 240 and the data is augmented with Gaussian noise ($\sigma = 0.01$) at each iteration. The model is trained for 45 epochs, then for a further 50 epochs with the learning rate reduced by a factor of 10.

\subsubsection{Results} \label{sec:emnist results}

\begin{figure}[ht] 
    \vspace{-5mm}
    \begin{center}
        \subfloat[Variable 1: upper width]{\includegraphics[width=0.25\textwidth]{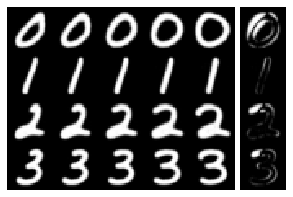}}
        \subfloat[Variable 8: lower width]{\includegraphics[width=0.25\textwidth]{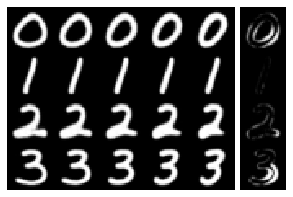}}
        \subfloat[Variable 3: height]{\includegraphics[width=0.25\textwidth]{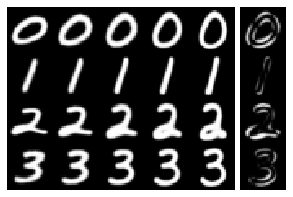}}
        \subfloat[Variable 4: bend]{\includegraphics[width=0.25\textwidth]{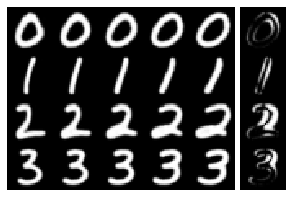}}
    \end{center}
    \vspace{-4mm}
    \caption{Selection of global latent variables found by GIN. Each row is conditioned on a different digit label. The variable runs from -2 to +2 standard deviations across the columns, with all other variables held constant at their mean value. The rightmost column shows a heatmap of the image areas most affected by the variable, computed as the absolute pixel difference between -1 and +1 standard deviations. Variable 1 controls the width of the top half of a digit, whereas variable 8 controls the width of the bottom half. Width in both cases is somewhat entangled with slant. Variable 3 controls the height and variable 4 controls how bent the digit is. Full set of variables for all digits in Appendix \ref{emnist figures}.}
    \label{fig:emnist global}
\end{figure}

\begin{figure}[ht] 
    \vspace{-5mm}
    \begin{center}
        \subfloat[Variable 11]{\includegraphics[width=0.25\textwidth]{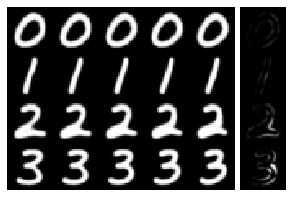}}
        \subfloat[Variable 12]{\includegraphics[width=0.25\textwidth]{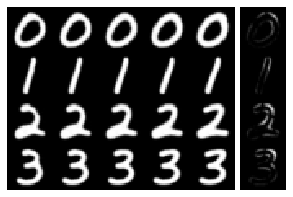}}
        \subfloat[Variable 13]{\includegraphics[width=0.25\textwidth]{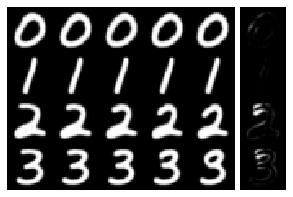}}
        \subfloat[Variable 16]{\includegraphics[width=0.25\textwidth]{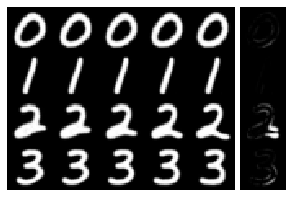}}
    \end{center}
    \vspace{-4mm}
    \caption{Selection of local latent variables found by GIN. Refer to Figure \ref{fig:emnist global} above for explanation of the layout. Variable 11 controls the shape of curved lines in the middle right. Digits without such lines are not affected. Variable 12 controls extension towards the upper right. Variable 13 modifies the top left of 2, 3 and 7 only (7 not shown here) and variable 16 modifies only the lower right stroke of a 2. Full set of variables for all digits in Appendix \ref{emnist figures}.}
    \label{fig:emnist local}
\end{figure}

The model encodes information into 22 broadly interpretable latent space variables. The eight variables with the highest standard deviation encode global information, such as slant, line thickness and width which affect all digits more or less equally. The remaining 14 meaningful variables encode more local information, which does not affect all digits equally. We interpret all other variables as encoding noise. This is motivated mainly by our observation of an abrupt end to any observable change in the reconstructed digits when any variable after the first 22 is changed. This can be seen clearly in Fig. \ref{fig:final variables} in Appendix \ref{emnist figures}.
Some variables, particularly the global ones, are not entirely disentangled from another. Nevertheless, the results are compelling and suggest that the assumptions required by the theory are approximately met.

We observed some global features which are not usually seen in disentanglement experiments on MNIST \citep[e.g.][]{chen2016infogan,dupont2018learning}. These experiments usually obtain digit slant, width and line thickness as the major global independent variables. We too observe slant and line thickness as independent variables (variables 2 and 5), but find width to be split into two variables, one governing the width of the upper half of a digit (variable 1) and the other the lower half (variable 8). This makes sense, since these can in fact vary independently. We also observe the height of a digit (variable 3) (see Fig. \ref{fig:emnist global}). This does not usually appear in disentanglement experiments, possibly because it is too subtle a variation for those experiments, but possibly because it is not present or not discoverable in the smaller MNIST dataset but is in EMNIST.

Local features are also usually not observed in such experiments, so the variables which control these are particularly interesting. These variables modify only a region of the digit, leaving the rest of the digit untouched. In addition, digits which do not have the feature which is being modified in that region are left alone. Examples include variable 13, which changes the orientation of the top-left stroke in 2, 3 and 7, and variable 16, which modifies only the lower-right stroke of a 2 (see Fig. \ref{fig:emnist local}). The full set of local and global variables can be seen in Fig. \ref{fig:first emnist variables} to Fig. \ref{fig:final variables} in Appendix \ref{emnist figures}.


\subsubsection{Note on the choice of conditioning variable}
The concept of a ``true generative process" rests on the assumption that conditioning variables $\ru$ act as causes for the latent variables $\rz$, such that the elements of the latter become independent given the former. 
In the case of handwritten digits, $\ru$ represents a person's decision to write a particular digit, and $\rz$ determines the hand motions to produce this digit. Modelling $\ru$ in terms of digit labels is a plausible proxy for the unknown causal variables in the human brain, and the success of the resulting model suggests that the assumptions are approximately fulfilled. Identifying promising conditioning variables $\ru$ in less obvious situations is an important open problem for future work.

\subsubsection{Potential Improvements}
Increasing the number of conditioning variables would bring this work closer in line with the existing theory. At the current value of ten, the theory only applies if there are at most four informative generating latent variables, which is almost certainly too low a number. One option to increase the number of conditions is to relax the labels from hard to soft, i.e.~compute posterior probabilities of class membership given a data example. This could be achieved by information distillation as in \citet{hinton2015distilling}. Another possibility is to split existing labels into sub-labels, for example making a distinction between sevens with and without crossbars. This may also aid the generation of more realistic samples, since we observed no sevens with crossbars in our generated samples, even though they make up approximately 5\% of the sevens in the dataset.

\section{Conclusion and Outlook}
We have expanded the theory of nonlinear ICA to cover problems with unknown intrinsic dimension and demonstrated an implementation with GIN, a new volume-preserving modification of the RealNVP invertible network architecture. The variables discovered by GIN in EMNIST are interpretable and detailed enough to suggest that the assumptions made about the generating process are approximately true. Furthermore, our experiments with EMNIST demonstrate the viability of applying models inspired by the new theory of nonlinear ICA to real-world data, even when not all of the conditions of the theory are met.

It is not clear if the methods from this work will scale to larger problems. However, given the recent advances of similar flow-based generative models in density estimation on larger datasets such as CelebA and ImageNet \citep[e.g.][]{kingma2018glow,ho2019flow++}, it is a plausible prospect. In addition, it is not clear whether the method can successfully be extended to the context of semi-supervised learning, or ultimately, unsupervised learning.

\clearpage
\bibliography{references}
\bibliographystyle{iclr2020_conference}

\clearpage
\section*{Supplementary Material}
\appendix
\section{Proof of identifiability} \label{appendix:proofs}
This section reproduces a proof from \citet{khemakhem2019variational} with some modifications to adapt it to the context of invertible neural networks (normalizing flows).

Define the domain of $f_{\varepsilon}$ as $\mathcal{Z} = \mathcal{Z}_1 \times \cdots \times \mathcal{Z}_n$. Define the vector of sufficient statistics $\mathbf{T}(\rz) = (T_{1,1}(z_1), \dots, T_{n,k}(z_n))$ and the vector of their coefficients $\bm{\lambda}(\ru) = (\lambda_{1,1}(\ru), \dots, \lambda_{n,k}(\ru))$.

\begin{theorem}
    Assume we observe data distributed according to the generative model defined in Sec. \ref{existing theory section}. Further suppose the following:
    \begin{enumerate}[label=(\roman*)]
        \item The sufficient statistics $T_{i,j}(z)$ are differentiable almost everywhere and their derivatives $\mathrm{d}T_{i,j}(z)/\mathrm{d}z$ are nonzero almost surely for all $z \in \mathcal{Z}_i$ and all $1 \leq i \leq n$ and $1 \leq j \leq k$.
        \item \label{cond:num conditions} There exist $nk+1$ distinct conditions $\ru_0, \dots, \ru_{nk}$ such that the matrix
        \begin{equation}
            \mL = (\bm{\lambda}(\ru_1) - \bm{\lambda}(\ru_0), \dots, \bm{\lambda}(\ru_{nk}) - \bm{\lambda}(\ru_0))
        \end{equation}
        of size $nk \times nk$ is invertible.
    \end{enumerate}
    Then the sufficient statistics of the generating latent space are related to those of the estimated latent space by the following relationship:
    \begin{equation}
        \mathbf{T}(\rz) = \mA \mathbf{T}'(\rw) + \vc
    \end{equation} 
    where $\mA$ is a constant, full-rank $nk \times dk$ matrix and $\vc \in \R^{nk}$ a constant vector.
\end{theorem}

\textbf{Proof:} 
The conditional probabilities $p_{\mathbf{T},\bm{\lambda},f,\varepsilon}(\rx|\ru)$ and $p_{\mathbf{T}',\bm{\lambda}',g}(\rx|\ru)$ are assumed to be the same in the limit of infinite data. By expanding these expressions via the change of variables formula and taking the logarithm we find
\begin{equation}
    \log p_{\mathbf{T},\bm{\lambda}}(\rz|\ru) + \log p(\varepsilon) + \log \left| \det J_{f^{-1}}(\rx) \right| = \log p_{\mathbf{T}',\bm{\lambda}'}(\rw|\ru) + \log \left| \det J_{g^{-1}}(\rx) \right|
\end{equation}
where $J$ is the Jacobian matrix. Let $\ru_0, \dots, \ru_{nk}$ be the conditions from \ref{cond:num conditions} above. We can subtract this expression for $\ru_0$ from the expression for some condition $\ru_l$. The Jacobian terms and the term involving $\varepsilon$ will vanish, since they do not depend on $\ru$:
\begin{equation}
    \log p_{\rz}(\rz|\ru_l) - \log p_{\rz}(\rz|\ru_0) = \log p_{\rw}(\rw|\ru_l) - \log p_{\rw}(\rw|\ru_0).
\end{equation}
Since the conditional probabilities of $\rz$ and $\rw$ are exponential family members, we can write this expression as
\begin{multline} \label{eqn:big zw}
    \sum_{i=1}^n \left[ \log \frac{Z_i(\ru_0)}{Z_i(\ru_l)} + \sum_{j=1}^k T_{i,j}(\rz) (\lambda_{i,j}(\ru_l) - \lambda_{i,j}(\ru_0)) \right] \\= \sum_{i=1}^d \left[ \log \frac{Z_i'(\ru_0)}{Z_i'(\ru_l)} + \sum_{j=1}^k T_{i,j}'(\rw) (\lambda_{i,j}'(\ru_l) - \lambda_{i,j}'(\ru_0)) \right]
\end{multline}
where the base measures $Q_i$ have cancelled out, since they do not depend on $\ru$. Defining $\bar{\bm{\lambda}}(\ru) = \bm{\lambda}(\ru) - \bm{\lambda}(\ru_0)$ and writing the above in terms of inner products, we find
\begin{equation}
    \left< \mathbf{\mT(\rz)}, \bar{\bm{\lambda}} \right> + \sum_i \log \frac{Z_i(\ru_0)}{Z_i(\ru_l)} = \left< \mathbf{\mT'(\rw)}, \bar{\bm{\lambda}}' \right> + \sum_i \log \frac{Z_i'(\ru_0)}{Z_i'(\ru_l)}
\end{equation}
for $1 \leq l \leq nk$. Combining the $nk$ expressions into a single matrix equation we can write this in terms of $\mL$ from condition \ref{cond:num conditions} above and an analogously defined $\mL' \in \R^{dk \times nk}$:
\begin{equation}
    \mL^T \mathbf{\mT(\rz)} = \mL'^T \mathbf{T'(\rw)} + \vb
\end{equation}
where $b_l = \sum_i \log \frac{Z_i'(\ru_0) Z_i(\ru_l)}{Z_i(\ru_0) Z_i'(\ru_l)}$. Since $\mL^T$ is invertible, we can multiply this expression by its inverse from the left to get
\begin{equation} \label{eqn:TATc}
    \mathbf{\mT(\rz)} = \mA \mathbf{T'(\rw)} + \vc
\end{equation}
where $\mA = \mL^{-T} \mL'^{T}$ and $\vc = \mL^{-T} \vb$. It now remains to show that $\mA$ has full rank.

According to a lemma from \citet{khemakhem2019variational}, there exist $k$ distinct values $z_i^1$ to $z_i^k$ such that $\left(\frac{\mathrm{d}T_i}{\mathrm{d}z_i}(z_i^1), \dots, \frac{\mathrm{d}T_i}{\mathrm{d}z_i}(z_i^k)\right)$ are linearly independent in $\R^k$, for all $1 \leq i \leq n$. Define $k$ vectors $\rz^l = (z_1^l, \dots, z_n^l)$ from the points given by this lemma. Take the derivative of equation (\ref{eqn:TATc}) evaluated at each of these $k$ vectors and concatentate the resulting Jacobians as $\mQ = [J_{\mathbf{T}}(\rz^1), \dots, J_{\mathbf{T}}(\rz^k)]$. Each Jacobian has size $nk \times n$, hence $\mQ$ has size $nk \times nk$ and is invertible by the lemma and the fact that each component of $\mathbf{T}$ is univariate. We can construct a corresponding matrix $\mQ'$ made up of the Jacobians of $\mathbf{T'}(g^{-1} \circ f_{\varepsilon}(\rz))$ evaluated at the same points and write
\begin{equation}
    \mQ = \mA \mQ'.
\end{equation}
Since $\mQ$ is full rank, both $\mA$ and $\mQ'$ must also be full rank.

\subsection{Remark regarding condition \ref{cond:num conditions}} \label{appendix:remark}
If the coefficients of the sufficient statistics $\bm{\lambda}$ are generated randomly and independently, then condition \ref{cond:num conditions} is almost surely fulfilled. In this case, we can ignore the dependence on $\ru$ to consider the $\bm{\lambda}(\ru_l)$ as independent random variables. Then condition \ref{cond:num conditions} states that instances of these random variables are in general position in $\R^{nk}$, which is true almost surely.

\subsection{Remark regarding noise variables in the estimated latent space}
Any variables in the estimated latent space whose distribution does not depend on the conditioning variable $\ru$ is considered as encoding only noise. Such variables will be cancelled out in equation (\ref{eqn:big zw}) and the corresponding column of $\mL'^T$ will contain only zeros. This means the corresponding column of $\mA$ will also contain only zeros, so any variation in such a noise variable has no effect on $\rz$. The reverse is also true: if a column of $\mA$ contains only zeros, the corresponding variable in the estimated latent space does not depend on $\ru$ and must encode only noise.



\section{Sparsity in unmixing matrix: Gaussian distribution} \label{gaussian general case}

Suppose samples $\rz$ from the generating latent space follow a conditional Gaussian distribution.
Suppose the estimating model $g$ faithfully reproduces the observed conditional density $p(\rx|\ru)$ and samples $\rw$ from its latent space also follow a conditional Gaussian distribution. Then we can apply equation (\ref{latent space relation}) to relate the generating and estimating latent spaces.
The sufficient statistics of a normal distribution with free mean and variance are $z$ and $z^2$. Hence the relationship between the latent spaces becomes
\begin{equation}
    \begin{pmatrix}
        \rz \\ \rz^2
    \end{pmatrix}
    = \mA \begin{pmatrix}
        \rw \\ \rw^2
    \end{pmatrix}
    + \vc
\end{equation}
where the squaring is applied element-wise. We can write $\mA$ in block matrix form as
\begin{equation}
    \mA = \begin{pmatrix}
        \mA^{(1)} & \mA^{(2)} \\
        \mA^{(3)} & \mA^{(4)}
    \end{pmatrix} 
\end{equation}
and $\vc$ as
\begin{equation}
    \vc = \begin{pmatrix}
        \vc^{(1)} \\ \vc^{(2)}
    \end{pmatrix}
\end{equation}
Then:
\begin{align}
    \rz &= \mA^{(1)} \rw + \mA^{(2)} \rw^2 + \vc^{(1)} \\
    \rz^2 &= \mA^{(3)} \rw + \mA^{(4)} \rw^2 + \vc^{(2)}
\end{align}
so we can write for each dimension $i$ of $\rz$
\begin{align}
    z_i &= \sum_j A^{(1)}_{ij} w_j + \sum_j A^{(2)}_{ij} w_j^2 + c^{(1)}_i \label{z long general} \\
    z_i^2 &= \sum_j A^{(3)}_{ij} w_j + \sum_j A^{(4)}_{ij} w_j^2 + c^{(2)}_i \label{z2 long general}
\end{align}
In order to compare the equations, we need to square (\ref{z long general}). To do so, we will have to square the second term on the right hand side, involving $w_j^2$. There is no matching term in (\ref{z2 long general}), so we have to set all entries of $\mA^{(2)}$ to zero. In more detail:
\begin{align}
    \left( \sum_j A^{(2)}_{ij} w_j^2 \right)^2 
        &= \sum_j \sum_{j'} A^{(2)}_{ij} A^{(2)}_{ij'} w_j^2 w_{j'}^2 \\
        &= \sum_j (A^{(2)}_{ij})^2 w_j^4 + \sum_{j \neq j'} A^{(2)}_{ij} A^{(2)}_{ij'} w_j^2 w_{j'}^2
\end{align}
The first term with $w_j^4$ matches no term in (\ref{z2 long general}), so we have to set $A^{(2)}_{ij}=0$ for all $i$ and $j$. This simplifies the earlier equation:
\begin{equation}
    z_i = \sum_j A^{(1)}_{ij} w_j + c^{(1)}_i
\end{equation}
The square of the first term on the right hand side involves terms with $w_j w_{j'}$ cross terms:
\begin{equation}
    \left( \sum_j A^{(1)}_{ij} w_j \right)^2 = \sum_j A^{(1)}_{ij} w_j^2 + \sum_{j \neq j'} A^{(1)}_{ij} A^{(1)}_{ij'} w_j w_{j'}
\end{equation}
so we have to set $\mA^{(1)}_{ij} \mA^{(1)}_{ij'} = 0$ for all $j \neq j'$. This means that the $i$-th row of $\mA^{(1)}$ can have at most one nonzero entry. It must also have at least one nonzero entry, since if the row were all zero, a row of $\mA$ would be all zero (since $\mA^{(2)} = 0$), but $\mA$ has full rank. Since there are as many or fewer rows than columns ($n \leq d$), each row of $\mA$ is linearly independent, so it is not possible for one to be zero. Hence each row of $\mA^{(1)}$ has exactly one nonzero entry. Moreover, no two rows of $\mA^{(1)}$ have their nonzero entries in the same column. If they did, the two rows would not be linearly independent, but they must be since $\mA$ has full rank. Therefore we can write
\begin{equation}
    z_i = a_i w_j + b_i
\end{equation}
where $a_i = A^{(1)}_{ij}$ and $b_i = c^{(1)}_i$. That is, the generating latent variable $z_i$ is linearly related to some latent variable of the estimating model $w_j$. This estimated latent variable is uniquely associated with $z_i$ and any estimated latent variables not associated with a generating latent variable $z_i$ (in the case $d > n$) encode no information about the generating latent space. So the model has decoded the original latent variables $\rz$ up to an affine transformation and permutation as a subset of variables in its estimated latent space and has encoded no information (only noise) into the remaining latent variables.

\section{Sparsity in unmixing matrix: two-parameter exponential family members} \label{two parameter general case}
The results of Appendix \ref{gaussian general case} can be extended to other members of the exponential family with 2 parameters. In the general case, writing $T_{i,1}=T_1$ and $T_{i,2}=T_2$ for all $i$, equations (\ref{z long general}) and (\ref{z2 long general}) become
\begin{align}
    T_1(z_i) &= \sum_j A^{(1)}_{ij} T_1(w_j) + \sum_j A^{(2)}_{ij} T_2(w_j) + c^{(1)}_i \label{eqn:t1} \\
    T_2(z_i) &= \sum_j A^{(3)}_{ij} T_1(w_j) + \sum_j A^{(4)}_{ij} T_2(w_j) + c^{(2)}_i 
\end{align}
which, with the definition $t = T_2 \circ T_1^{-1}$ becomes
\begin{align}
    T_1(z_i) &= \sum_j A^{(1)}_{ij} T_1(w_j) + \sum_j A^{(2)}_{ij} t(T_1(w_j)) + c^{(1)}_i \\
    t(T_1(w_j)) &= \sum_j A^{(3)}_{ij} T_1(w_j) + \sum_j A^{(4)}_{ij} t(T_1(w_j)) + c^{(2)}_i .
\end{align}
We can combine these equations to get
\begin{equation}
    t \left( \sum_j A^{(1)}_{ij} T_1(w_j) + \sum_j A^{(2)}_{ij} t(T_1(w_j)) + c^{(1)}_i \right) = \sum_j A^{(3)}_{ij} T_1(w_j) + \sum_j A^{(4)}_{ij} t(T_1(w_j)) + c^{(2)}_i.
\end{equation}
This equation has many summed terms on the right. Suppose that $t$ has a convergent Taylor expansion in some region of its domain. We can take this expansion of the term on the left and compare coefficients. Since $t$ cannot be linear,
there will be terms in the expansion of order two or higher. As in the Gaussian case, these polynomial terms create cross terms which are impossible to reconcile with those on the right hand side of the equation. The only consistent solutions can be found by setting all coefficients except those of functions of $w_j$ (for some $j$) to zero:
\begin{equation} \label{eqn:w_j relationship}
    t \left( A^{(1)}_{ij} T_1(w_j) + A^{(2)}_{ij} t(T_1(w_j)) + c^{(1)}_i \right) = A^{(3)}_{ij} T_1(w_j) + A^{(4)}_{ij} t(T_1(w_{j})) + c^{(2)}_i
\end{equation}
We can further simplify this equation by examining higher-order terms, but need to know the form of $t$ to do so. In any case, we can now write equation (\ref{eqn:t1}) as
\begin{equation}
    T_1(z_i) = A^{(1)}_{ij} T_1(w_j) + A^{(2)}_{ij} t(T_1(w_j)) + c^{(1)}_i
\end{equation}
showing that each generating latent variable $z_i$ is related to exactly one estimated latent variable $w_j$. As in the Gaussian case, we can use the full rank property of $\mA$ to see that each estimated latent variable is associated with at most one generating latent variable, and any estimated latent variables not associated with any generating latent variables must encode only noise.
The task now is to check the form of $t$ for each two-parameter member of the exponential family, to see what further constraints we can derive from equation (\ref{eqn:w_j relationship}). 
The results are stated in Table \ref{table:exponential family}.

\newcommand\T{\rule{0pt}{2.6ex}}       
\newcommand\B{\rule[-1.2ex]{0pt}{0pt}} 

\begin{table}[ht]
    \caption{Two-parameter exponential family members and selected properties.}
    \label{table:exponential family}
    \centering
    \begin{tabular}{|c|c|c|c|}
        \hline
        Distribution & Sufficient Statistics & $t(x)$ & Latent Space Relationship \T\B \\
        \hline
        Normal & $(z, z^2)$ & $x^2$ & $z_i = a w_j + c$ \T \\
        Lognormal & $(\log z, (\log z)^2)$ & $x^2$ & $\log z_i = a \log w_j + c$ \T \\
        Inverse Gaussian & $(z, 1/z)$ & $1/x$ & $z_i = a w_j$ or $z_i = a/w_j$ \T \\
        Gamma & $(\log z, z)$ & $\exp(x)$ & $z_i = a w_j$ \T \\
        Inverse Gamma & $(\log z, 1/z)$ & $\exp(-x)$ & $z_i = a w_j$ \T\B \\
        Beta & $(\log z, \log(1-z))$ & $\log(1-\exp(x))$ & \makecell{$\log z_i = \log w_j$\\or $\log z_i = \log(1-w_j)$} \T\B \\
        \hline
    \end{tabular}
\end{table}
\begin{center}
    
\end{center}

\clearpage
\section{Network Architecture} \label{inn architecture} 
The estimating model $g$ is built in the reverse direction for practical purposes, so the models described here are $g^{-1}$ which maps from the data space to the latent space. The type and number of coupling blocks for the different experiments are shown below. The affine coupling function is the concatentation of the scale function $s$ and the translation function $t$, computed together for efficiency, as in \citet{kingma2018glow}. It is implemented as either a fully connected network (MLP) or convolutional network, with the specified layer widths and a ReLU activation after all but the final layers. For the convolutional coupling blocks, the splits are along the channel dimension. The scale function $s$ is passed through a clamping function $2\tanh(s)$, which limits the output to the range (-2, 2), as in \citet{ardizzone2019guided}. Two affine coupling functions are applied per block, as described in \citet{dinh2016density}. Downsampling increases the number of channels by a factor of 4 and decreases the image width and height by a factor of 2, done in a checkerboard-like manner, as described in \citet{jacobsen2018revnet}. The dimensions are permuted in a random but fixed way before application of each fully connected coupling block and likewise the channels for the convolutional coupling blocks. The network for the artificial data experiments has 4,480 learnable parameters and the network for the EMNIST experiments has 2,620,192 learnable parameters.

\begin{table}[h]
    \caption{Network architecture for artificial data experiments}
    \label{table:artificial data}
    \centering
    \begin{tabular}{|c|c|c|c|}
        \hline
        Type of block & Number & Input shape & Affine coupling function layer widths \T\B \\
        \hline
        Fully Connected Coupling & 8 & 10 & $5 \rightarrow 10 \rightarrow 10 \rightarrow 10$ \T\B \\
        \hline
    \end{tabular}
\end{table}

\begin{table}[h]
    \caption{Network architecture for EMNIST experiments}
    \label{table:emnist}
    \centering
    \begin{tabular}{|c|c|c|c|}
        \hline
        Type of block & Number & Input shape & Affine coupling function layer widths \T\B \\
        \hline
        Downsampling & 1 & (1, 28, 28) & \T \\
        Convolutional Coupling & 4 & (4, 14, 14) & 2 $\rightarrow 16 \rightarrow 16 \rightarrow 4$ \T \\
        Downsampling & 1 & (4, 14, 14) & \T \\
        Convolutional Coupling & 4 & (16, 7, 7) & 8 $\rightarrow 32 \rightarrow 32 \rightarrow 16$ \T \\
        Flattening & 1 & (16, 7, 7) & \T \\
        Fully Connected Coupling & 2 & 784 & 392 $\rightarrow 392 \rightarrow 392 \rightarrow 784$ \T\B \\
        \hline
    \end{tabular}
\end{table}

\subsection{Note on Optimization Method}
In the experiments described in this paper, the mean and variance of a mixture component was updated at each iteration as the mean and variance of the transformations to latent space of all data points belonging to that mixture component in a minibatch of data $\mathcal{D}$:
\begin{align}
    \mu_i(\ru'; \vtheta) &\leftarrow \E_{\mathcal{D}: \ru=\ru'} (g_i^{-1}(\rx;\vtheta)) \\
    \sigma_i^2(\ru'; \vtheta) &\leftarrow \text{Var}_{\mathcal{D}: \ru=\ru'}(g_i^{-1}(\rx;\vtheta)).
\end{align}
Hence the parameters of the mixture components would change in each batch, according to the data present. The notation $\mu_i(\ru; \vtheta)$ does not indicate that $\mu$ is directly parameterized by $\vtheta$ and learned, instead it indicates that $\mu$ is a function of $g^{-1}$ which is parameterized by $\vtheta$. A change in $\vtheta$ will also change the output of $\mu$, given the same mini-batch of data $\mathcal{D}$. The same holds for $\sigma_i(\ru; \vtheta)$.

We expect that specifying the means and variances as learnable parameters updated in tandem with the model weights would have worked equally well.

\clearpage
\section{Figures from the Artificial Data Experiments} \label{appendix:artificial figs}

\begin{figure}[h] 
    \begin{center}
        \includegraphics[width=0.9\textwidth]{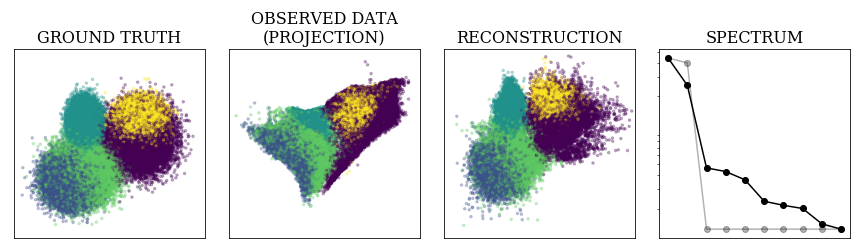}
    \end{center}
    \caption{Experiment 1: Five mixture components and 100,000 data points. GIN successfully reconstructs the generating latent space and gives importance to only two of its ten latent variables, reflecting the two-dimensional nature of the generating latent space.
    }
    \label{5 clusters large dataset fig}
\end{figure}

\begin{figure}[h] 
    \begin{center}
        \includegraphics[width=0.9\textwidth]{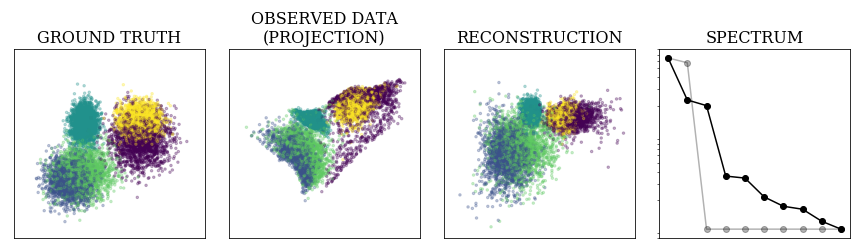}
    \end{center}
    \caption{Same experiment as in Figure \ref{5 clusters large dataset fig} with the number of data points reduced to 10,000. GIN fails to successfully estimate the ground truth latent variables, due to limited data: The first variable ($x$-axis) is well approximated in the reconstruction, but the second variable ($y$-axis) is split into two in the reconstruction, one capturing mainly information about the lower two clusters (shown here) and another information about the other mixture components (not shown). We also observe a less clear spectrum, where three variables are given more importance than the rest, not faithfully reflecting the two-dimensional nature of the generating latent space.}
    \label{5 clusters small dataset fig}
\end{figure}

\begin{figure}[h] 
    \begin{center}
        \includegraphics[width=0.9\textwidth]{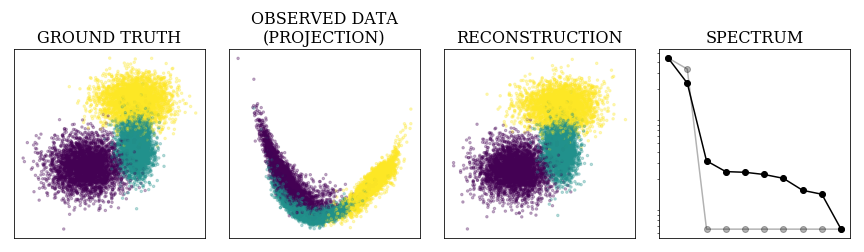}
    \end{center}
    \caption{Experiment 2: Only three mixture components (not sufficient for identifiability according to the theory). Nevertheless, GIN successfully reconstructs the ground truth latent variables. This suggests that the current theory of nonlinear ICA relies on sufficient, but not necessary, conditions for identifiability.}
    \label{3 clusters fig}
\end{figure}

\clearpage
\section{EMNIST Figures} \label{emnist figures}

\begin{figure}[h] 
    \begin{center}
        \subfloat[Full Temperature]{\includegraphics[width=0.5\textwidth]{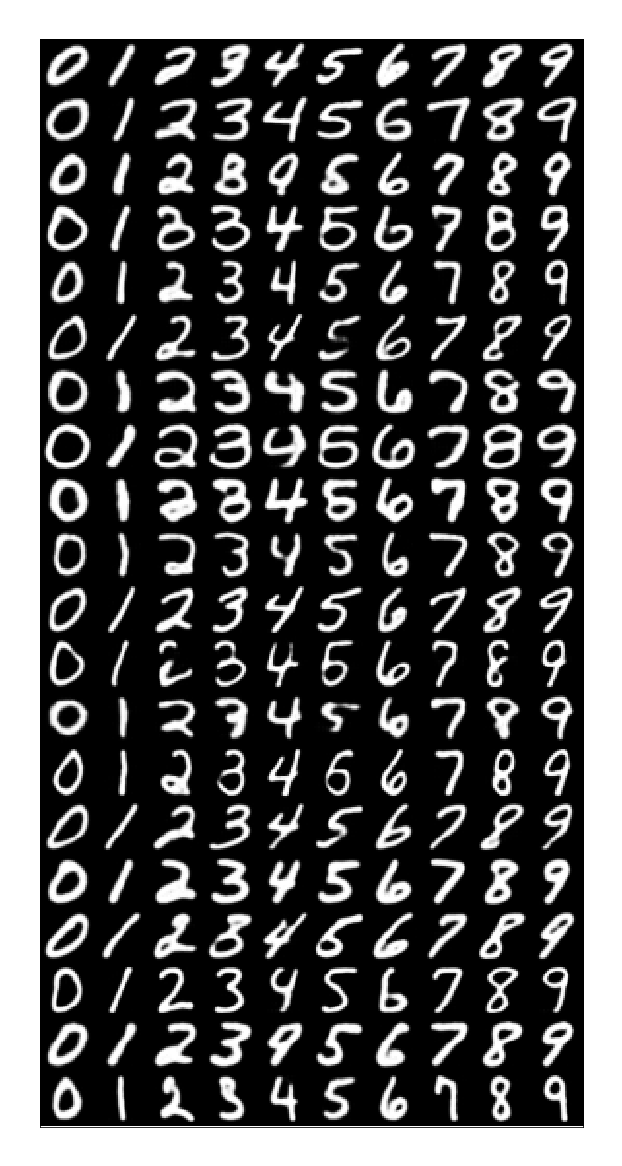}}
        \subfloat[Reduced Temperature (T = 0.8)]{\includegraphics[width=0.5\textwidth]{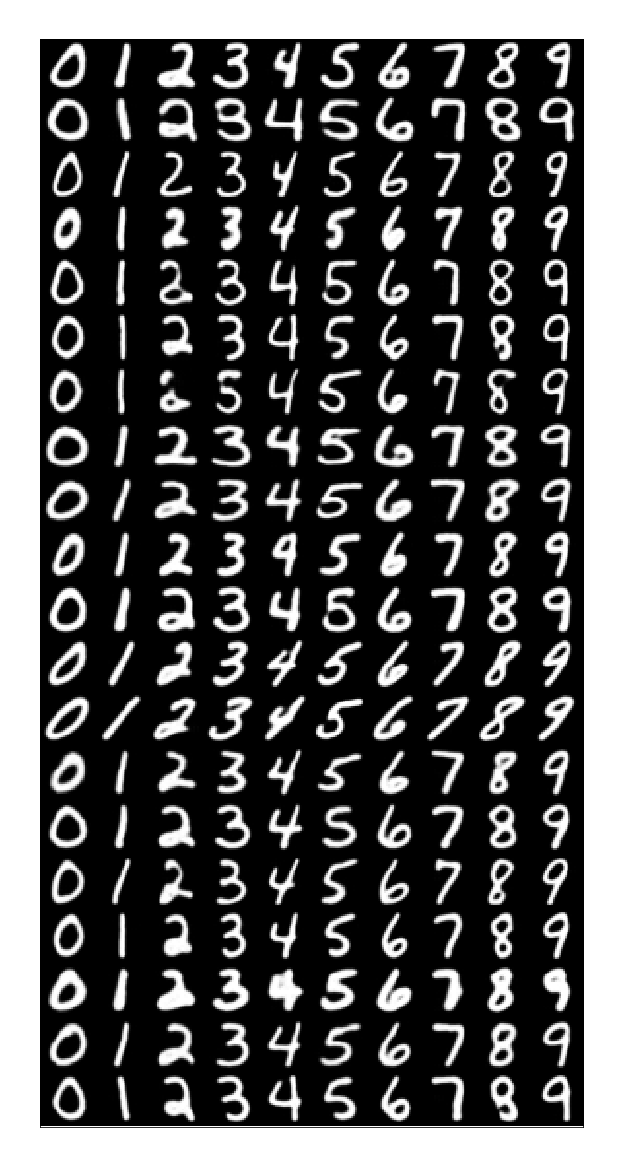}}
    \end{center}
    \caption{Full and reduced temperature samples from the model trained on EMNIST. Reduced temperature samples are made by sampling from a Gaussian distribution where the standard deviation is reduced by the temperature factor. The 22 most significant variables are sampled, with the others kept to their mean value. This eliminates noise from the images but preserves the full variability of digit shapes. Each row has the same latent code (whitened value) but is conditioned on a different class in each column, hence the style of the digits is consistent across rows.}
\end{figure}

\begin{figure}[h] 
    \begin{center}
        \subfloat[Variable 1: width of top half]{\includegraphics[width=0.5\textwidth]{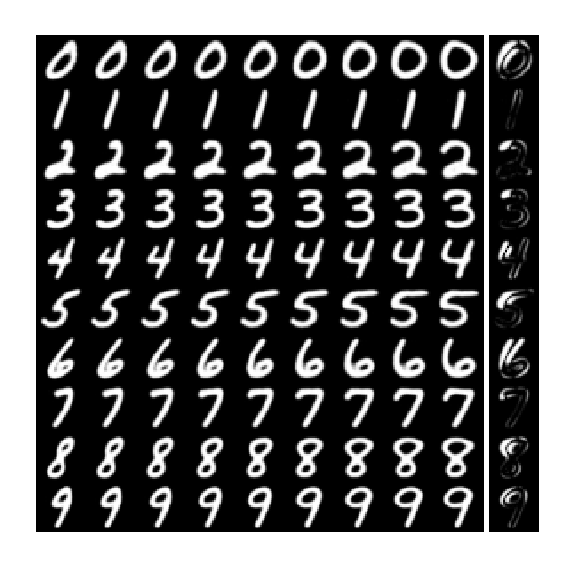}}
        \subfloat[Variable 2: slant/angle]{\includegraphics[width=0.5\textwidth]{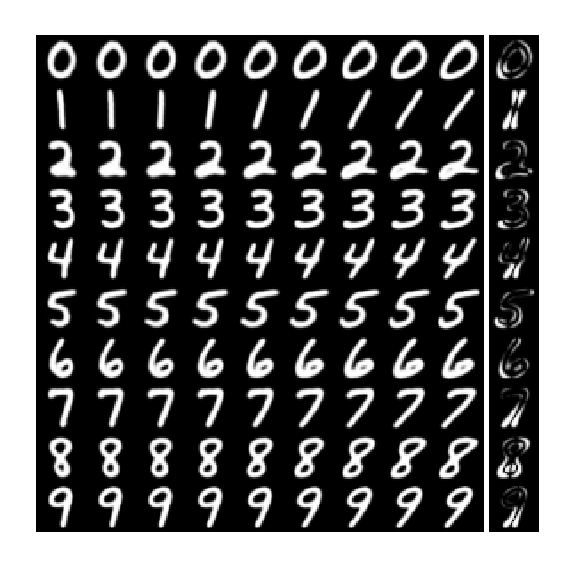}}
    \end{center}
    \begin{center}
        \subfloat[Variable 3: height]{\includegraphics[width=0.5\textwidth]{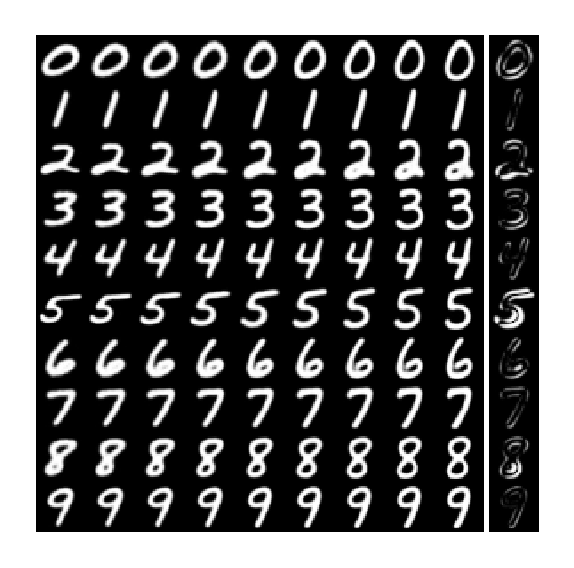}}
        \subfloat[Variable 4: bend through center]{\includegraphics[width=0.5\textwidth]{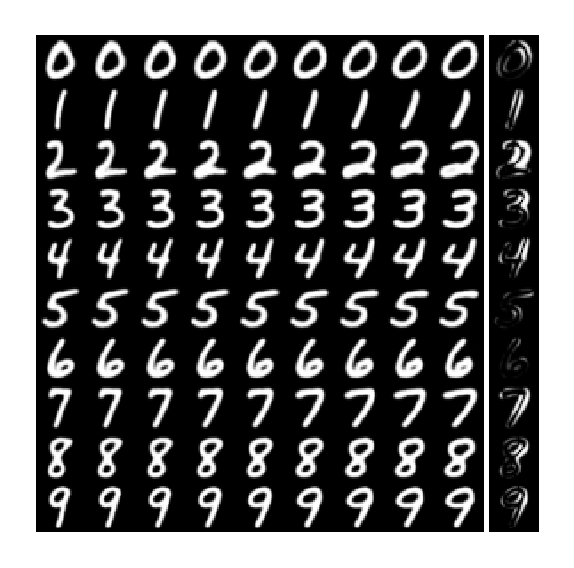}}
    \end{center}
    \caption{Most significant latent variables 1 to 4. Each row is conditioned on a different digit label. The variable runs from -2 to +2 standard deviations across the columns, with all other variables held constant at their mean value. The rightmost column shows a heatmap of the image areas most affected by the variable, computed as the absolute pixel difference between -1 and +1 standard deviations.}
    \label{fig:first emnist variables}
\end{figure}

\begin{figure}[h] 
    \begin{center}
        \subfloat[Variable 5: line thickness]{\includegraphics[width=0.5\textwidth]{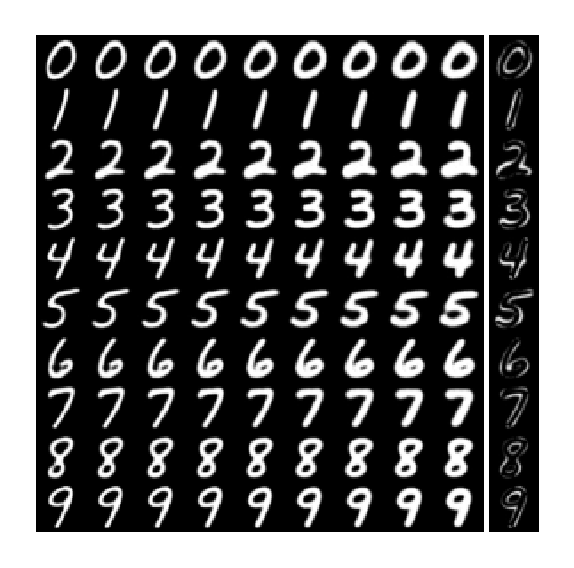}}
        \subfloat[Variable 6: slant of vertical bar in 4,7 and 9]{\includegraphics[width=0.5\textwidth]{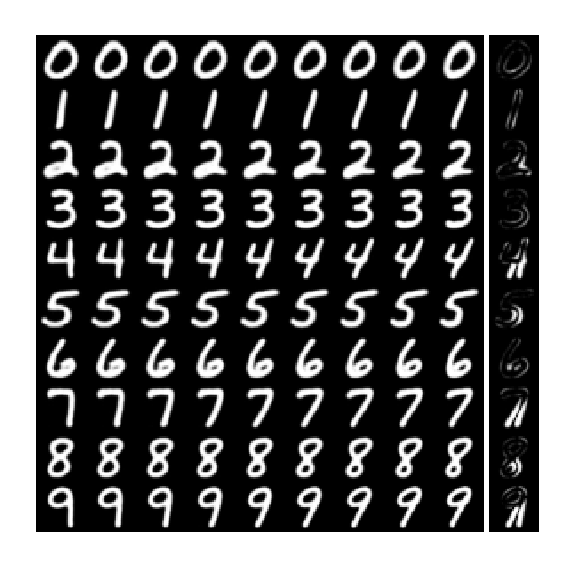}}
    \end{center}
    \begin{center}
        \subfloat[Variable 7: height of horizontal bar]{\includegraphics[width=0.5\textwidth]{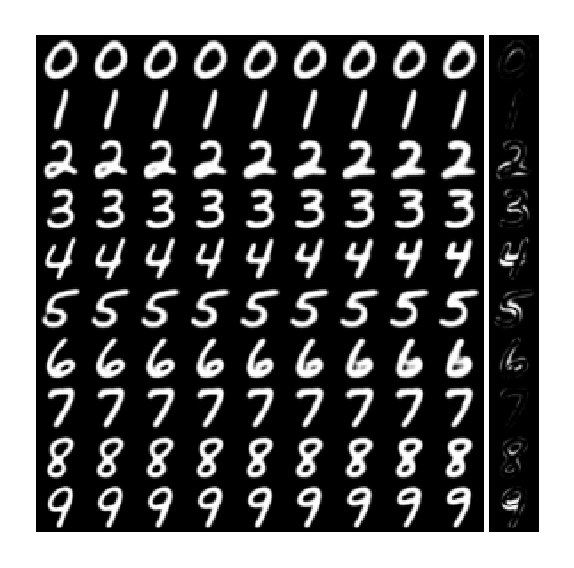}}
        \subfloat[Variable 8: width of bottom half]{\includegraphics[width=0.5\textwidth]{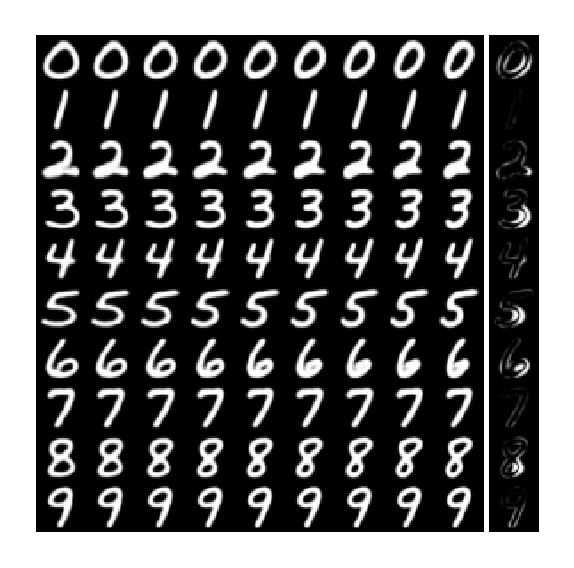}}
    \end{center}
    \caption{Most significant latent variables 5 to 8. Each row is conditioned on a different digit label. The variable runs from -2 to +2 standard deviations across the columns, with all other variables held constant at their mean value. The rightmost column shows a heatmap of the image areas most affected by the variable, computed as the absolute pixel difference between -1 and +1 standard deviations.}
\end{figure}

\begin{figure}[h] 
    \begin{center}
        \subfloat[Variable 9: extension of center left feature towards the left]{\includegraphics[width=0.5\textwidth]{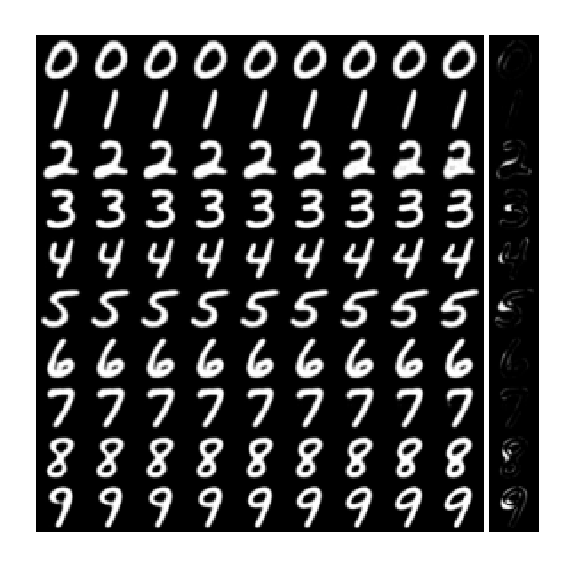}}
        \subfloat[Variable 10: openness of lower loop]{\includegraphics[width=0.5\textwidth]{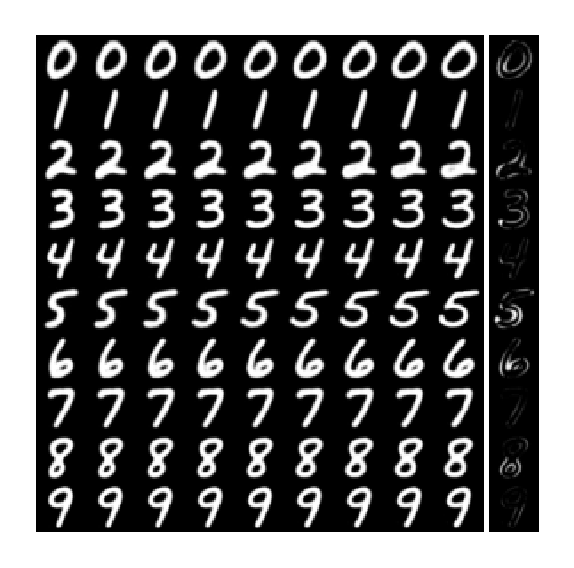}}
    \end{center}
    \begin{center}
        \subfloat[Variable 11: shape of center right feature]{\includegraphics[width=0.5\textwidth]{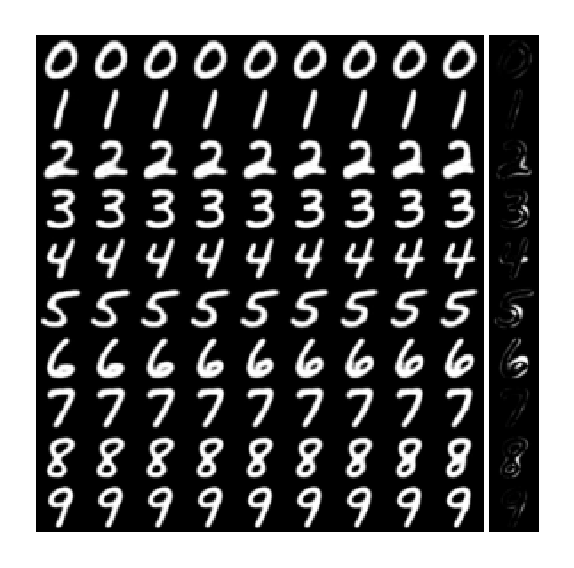}}
        \subfloat[Variable 12: top right corner]{\includegraphics[width=0.5\textwidth]{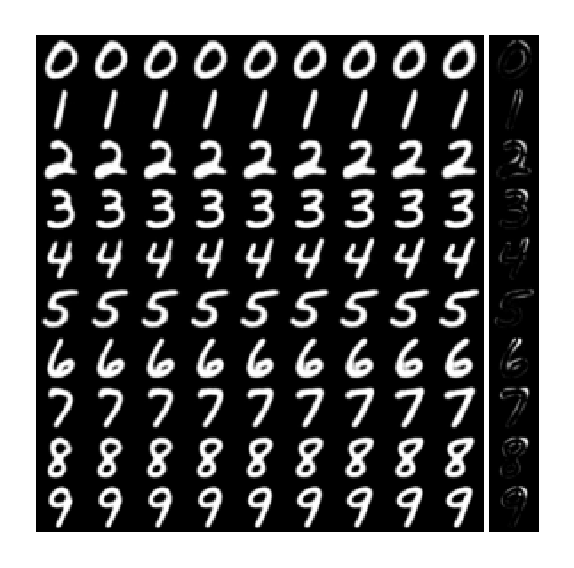}}
    \end{center}
    \caption{Most significant latent variables 9 to 12. Each row is conditioned on a different digit label. The variable runs from -2 to +2 standard deviations across the columns, with all other variables held constant at their mean value. The rightmost column shows a heatmap of the image areas most affected by the variable, computed as the absolute pixel difference between -1 and +1 standard deviations.}
\end{figure}

\begin{figure}[h] 
    \begin{center}
        \subfloat[Variable 13: orientation of stroke in top left corner for 2, 3 and 7]{\includegraphics[width=0.5\textwidth]{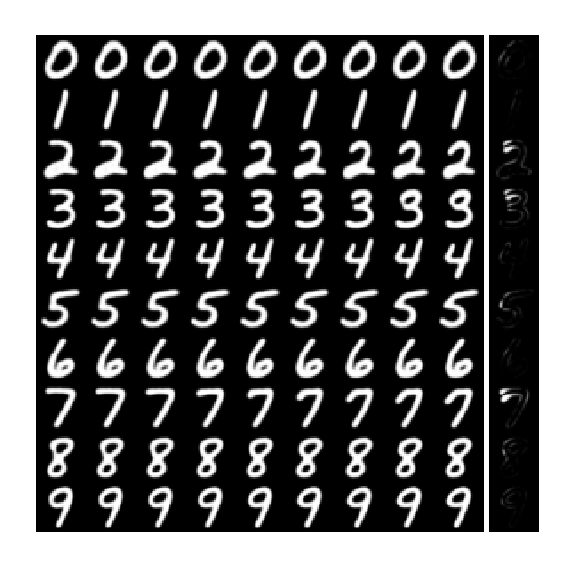}}
        \subfloat[Variable 14: shape of top part of bottom loop]{\includegraphics[width=0.5\textwidth]{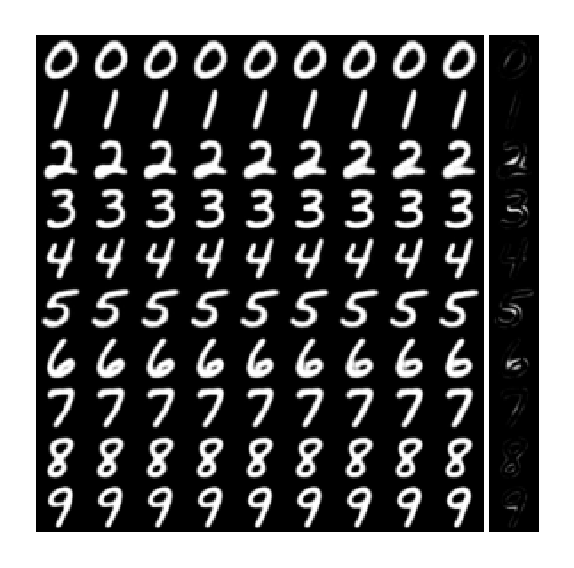}}
    \end{center}
    \begin{center}
        \subfloat[Variable 15: angle of centrally located horizontal features]{\includegraphics[width=0.5\textwidth]{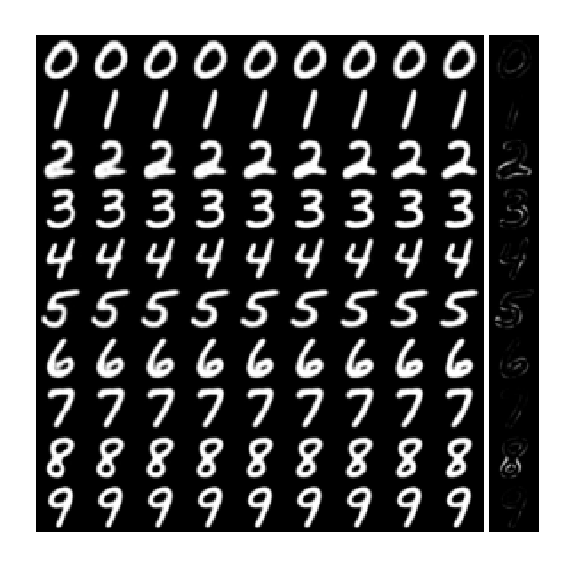}}
        \subfloat[Variable 16: bottom right stroke of 2]{\includegraphics[width=0.5\textwidth]{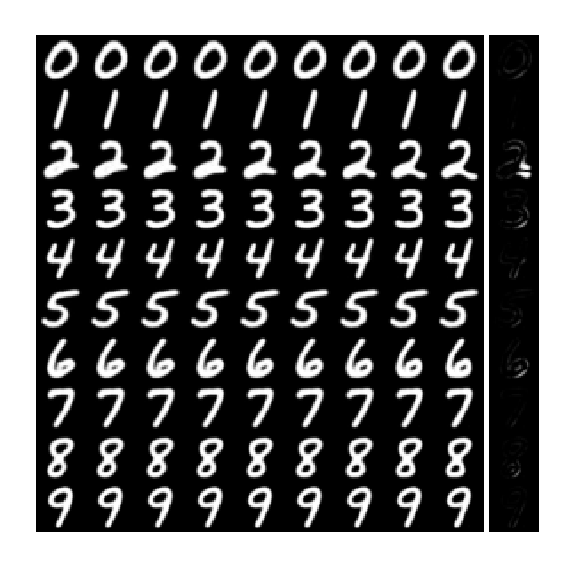}}
    \end{center}
    \caption{Most significant latent variables 13 to 16. Each row is conditioned on a different digit label. The variable runs from -2 to +2 standard deviations across the columns, with all other variables held constant at their mean value. The rightmost column shows a heatmap of the image areas most affected by the variable, computed as the absolute pixel difference between -1 and +1 standard deviations.}
\end{figure}

\begin{figure}[h] 
    \begin{center}
        \subfloat[Variable 17: top left stroke of 4]{\includegraphics[width=0.5\textwidth]{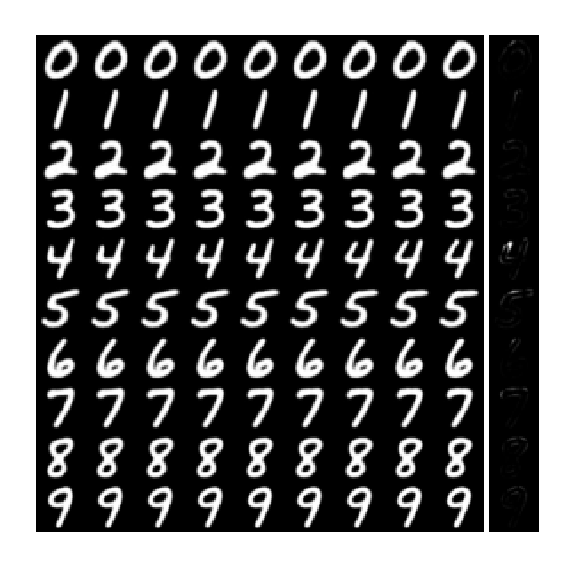}}
        \subfloat[Variable 18: top stroke of 5 and 7]{\includegraphics[width=0.5\textwidth]{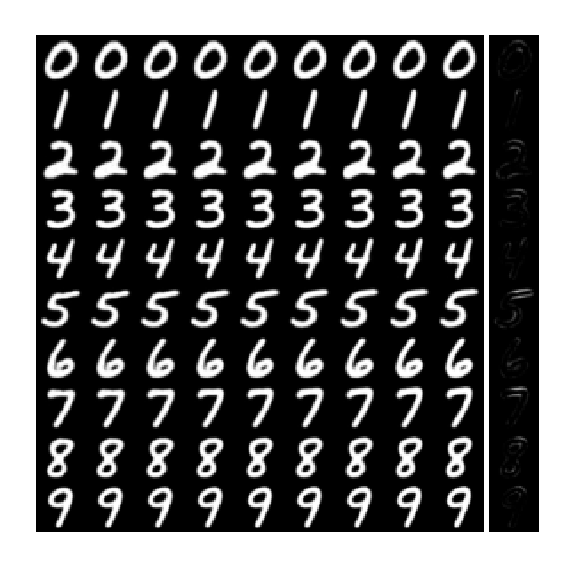}}
    \end{center}
    \begin{center}
        \subfloat[Variable 19: extension towards top right]{\includegraphics[width=0.5\textwidth]{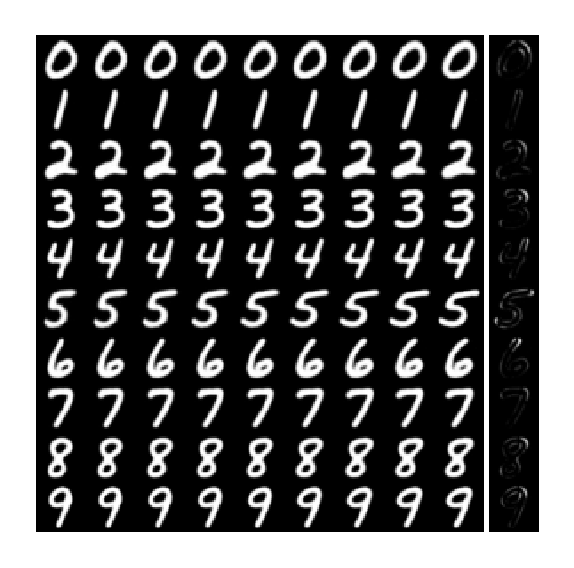}}
        \subfloat[Variable 20: curvature of vertical stroke of 4]{\includegraphics[width=0.5\textwidth]{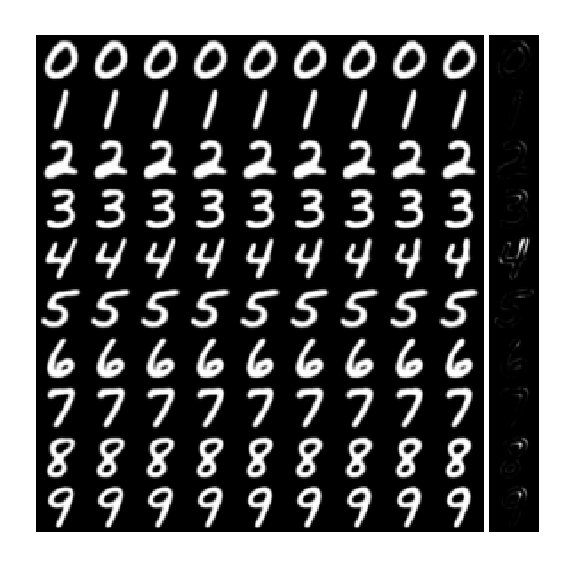}}
    \end{center}
    \caption{Most significant latent variables 17 to 20. Each row is conditioned on a different digit label. The variable runs from -2 to +2 standard deviations across the columns, with all other variables held constant at their mean value. The rightmost column shows a heatmap of the image areas most affected by the variable, computed as the absolute pixel difference between -1 and +1 standard deviations.}
\end{figure}

\begin{figure}[h] 
    \begin{center}
        \subfloat[Variable 21: thickness of upper loop]{\includegraphics[width=0.5\textwidth]{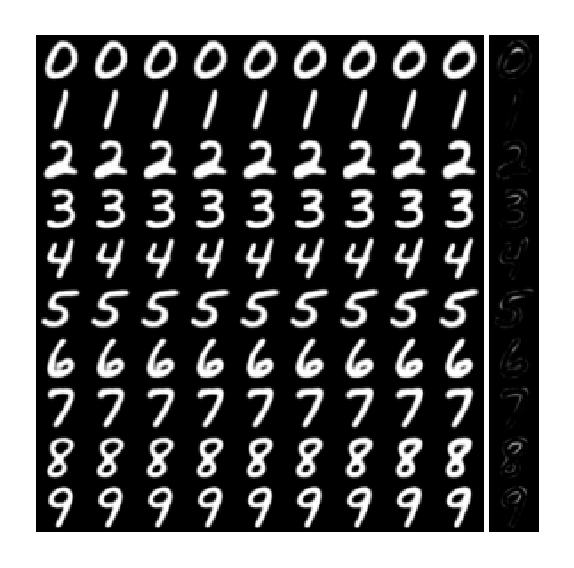}}
        \subfloat[Variable 22: extension towards top left]{\includegraphics[width=0.5\textwidth]{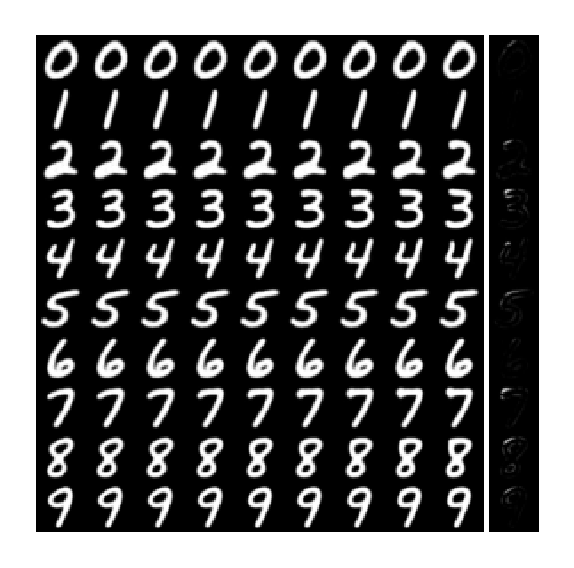}}
    \end{center}
    \begin{center}
        \subfloat[Variable 23: no effect]{\includegraphics[width=0.5\textwidth]{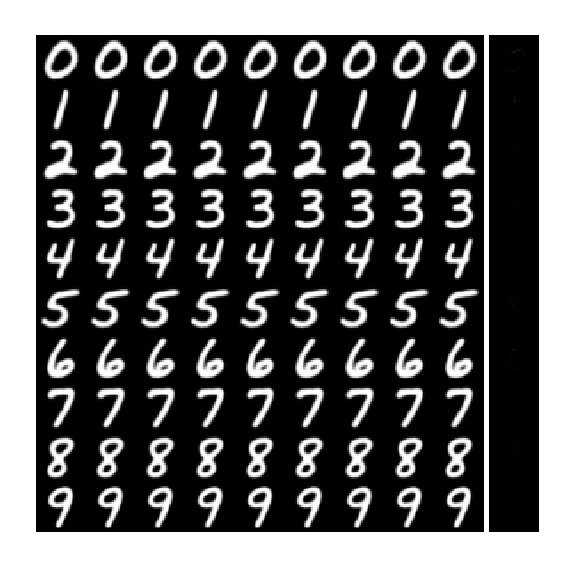}}
        \subfloat[Variable 24: no effect]{\includegraphics[width=0.5\textwidth]{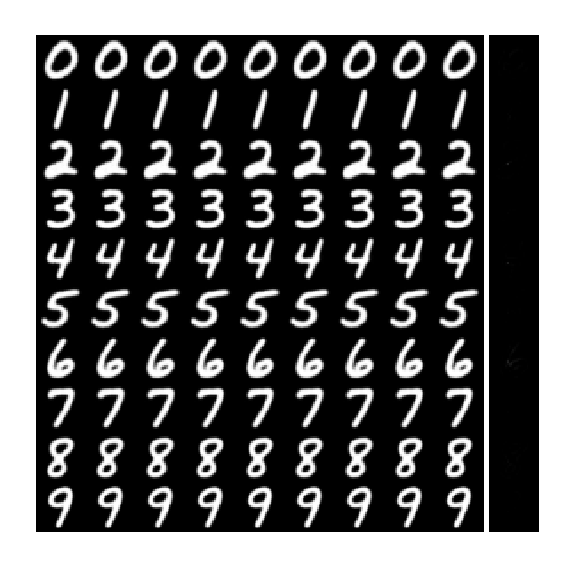}}
    \end{center}
    \caption{Most significant latent variables 21 to 24. Each row is conditioned on a different digit label. The variable runs from -2 to +2 standard deviations across the columns, with all other variables held constant at their mean value. The rightmost column shows a heatmap of the image areas most affected by the variable, computed as the absolute pixel difference between -1 and +1 standard deviations.}
    \label{fig:final variables}
\end{figure}

\end{document}